\PassOptionsToPackage{table}{xcolor}
\documentclass[11pt]{article}

\usepackage[preprint]{acl}
\usepackage{times}
\usepackage{latexsym}
\usepackage{amsmath}
\usepackage{amssymb}
\usepackage[T1]{fontenc}
\usepackage[utf8]{inputenc}
\usepackage{microtype}
\usepackage{inconsolata}
\usepackage{graphicx}
\usepackage{xcolor}
\usepackage{booktabs}
\usepackage{collcell}
\usepackage{xfp}
\usepackage{float}

\newcommand{\colorpos}[1]{%
  \if\relax\detokenize{#1}\relax
  \else
    \cellcolor{blue!\fpeval{round(#1*30,0)}!white}#1%
  \fi
}
\newcolumntype{C}{>{\collectcell\colorpos}r<{\endcollectcell}}
\newcommand{\xhdr}[1]{\vspace{1mm}\noindent{{\bf #1.}}}
\title{Categorize Early, Integrate Late:\\Divergent Processing Strategies in Automatic Speech Recognition}
\author{
  Nathan Roll\thanks{Equal contribution.}, 
  Pranav Bhalerao\footnotemark[1],
  Martijn Bartelds,
  Arjun Pawar,
  Yuka Tatsumi, \\
  \textbf{Tolulope Ogunremi,
  Chen Shani,
  Calbert Graham,
  Meghan Sumner, 
  Dan Jurafsky} \\[1ex]
  Stanford University
}

\begin{document}
\maketitle
\begin{abstract}
In speech language modeling, two architectures dominate the frontier: the Transformer and the Conformer. However, it remains unknown whether their comparable performance stems from convergent processing strategies or distinct architectural inductive biases. We introduce \textit{Architectural Fingerprinting}, a probing framework that isolates the effect of architecture on representation, and apply it to a controlled suite of 24 pre-trained encoders (39M--3.3B parameters). Our analysis reveals divergent hierarchies: Conformers implement a ``Categorize Early'' strategy, resolving phoneme categories 29\% earlier in depth and speaker gender by 16\% depth. In contrast, Transformers ``Integrate Late,'' deferring phoneme, accent, and duration encoding to deep layers (49-57\%). These fingerprints suggest design heuristics: Conformers' front-loaded categorization may benefit low-latency streaming, while Transformers' deep integration may favor tasks requiring rich context and cross-utterance normalization.
\end{abstract}

\section{Introduction}

In text modeling, architectural diversity has largely collapsed onto the Transformer paradigm. Speech is different: two encoder families, Transformers and Conformers, coexist at the frontier, achieving comparable word error rates (WER) on standard benchmarks \cite{gulati2020conformer, radford2023robust}. This coexistence creates a unique scientific opportunity. Unlike text, where representational convergence cannot be disentangled from architectural uniformity, speech allows us to ask: \textbf{when two architectures solve the same problem equally well, do they converge on similar internal processing strategies, or does architectural inductive bias produce fundamentally different solutions?}

This question matters for three reasons. First, \textbf{interpretability}: knowing where phonetic and speaker information emerges enables targeted debugging and trust calibration. Second, \textbf{demographic encoding}: understanding when gender and accent cues become accessible informs fairness audits. Third, \textbf{deployment}: if architectures process information at different depths, this has implications for streaming, latency, and layer-wise pruning.

Prior probing work has established that speech models encode acoustic and linguistic information hierarchically \cite{chrupala2017representations, belinkov2019analyzing, pasad2021layer, martin_probing_2023, pasad_comparative_2023}, with phonetic information peaking in middle layers and word identity emerging later \cite{pasad2021layer, BARTELDS2022101137}. Early interpretability work on speech representations includes \citet{chrupala2017representations}, which analyzes layers of a visually grounded speech model to study how linguistic structure emerges across depth. Complementary analyses have revealed speaker-specific and L1-related structure \cite{ferragne2019towards, graham2021l1}. Beyond speech, layer-wise model-comparison work has shown that architectures with similar end-task performance can nevertheless learn systematically different internal representations; for example, \citet{raghu2021dovisiontransformers} compare ViTs and CNNs and show clear differences in representation geometry across layers. However, no prior work systematically compares Transformer versus Conformer encoders to isolate how architectural inductive bias (rather than scale or training data) shapes representational hierarchies in ASR.

\begin{figure*}[t]
    \centering
    \includegraphics[width=\textwidth]{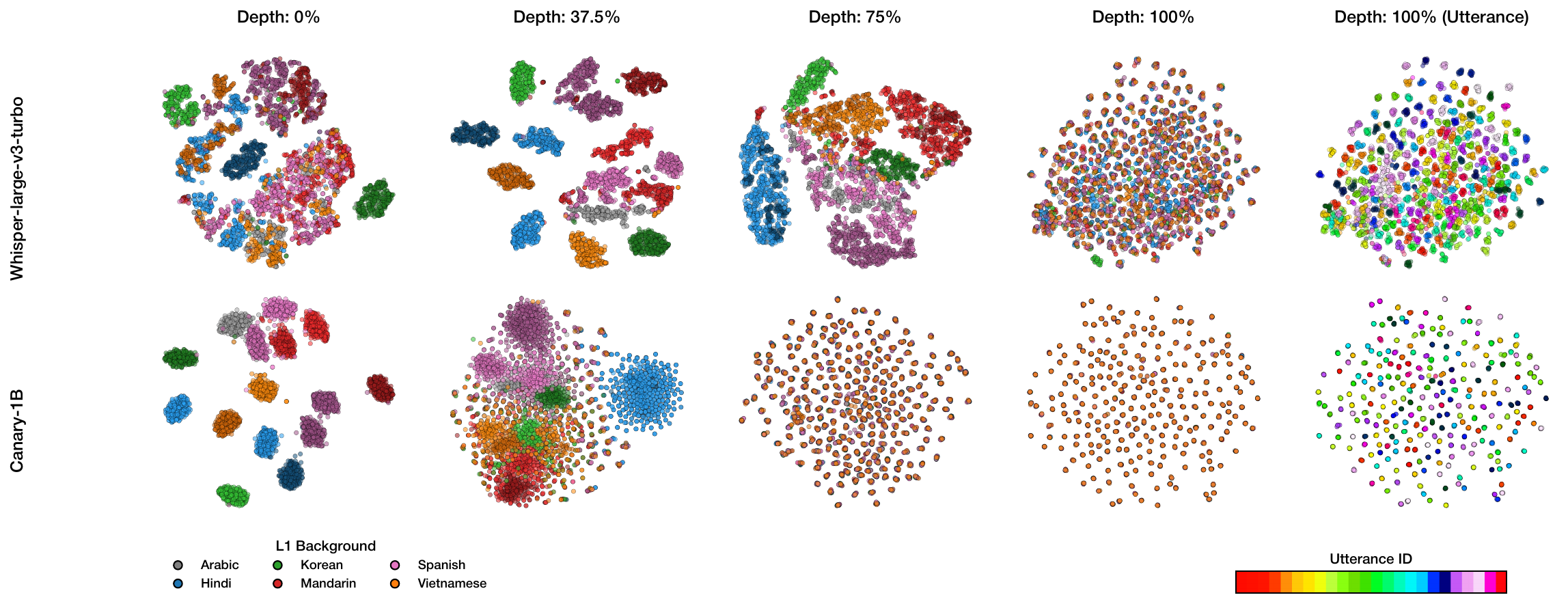}
    \caption{t-SNE visualization of encoder representations comparing a Transformer (Whisper-large-v3-turbo, top row) and a Conformer (Canary-1B, bottom row) across network depth, using samples from the L2-ARCTIC dataset. Each column shows representations at increasing depths (0\%, 37.5\%, 75\%, and 100\% of total layers). The first three columns are colored by L1 background: Arabic (gray), Hindi (blue), Korean (green), Mandarin (red), Spanish (magenta), and Vietnamese (orange). The rightmost column (100\% depth) uses a distinct color palette, colored by utterance identity, to reveal how content clusters in the final layer independent of L1 background. The legend displays a color gradient representing utterance identity. Utterance colors are assigned consistently across samples but do not correspond to specific L1 backgrounds. Points are plotted in random order within each depth column to avoid visual occlusion bias. The Transformer shows gradual emergence of accent clustering in deep layers, while the Conformer exhibits earlier and sharper accent separation, consistent with their divergent processing hierarchies.}
\label{fig:tsne}
\end{figure*}

Our main contributions are:
\begin{enumerate}
    \item We introduce \textit{Architectural Fingerprinting}, a probing framework using linear probes to quantify when features become linearly accessible across depth, characterizing how architectural inductive bias shapes representational hierarchies. Our probes measure linear accessibility rather than making causal claims about mechanism.
    \item We present a comprehensive, architecture-controlled analysis of 24 Transformer and Conformer models, revealing consistent, architecture-dependent representational profiles that persist across scale, dataset, and training regime despite comparable automatic speech recognition (ASR) performance.
    \item We identify two divergent processing strategies: Conformers' \textit{Categorize Early} strategy, which front-loads phoneme category and gender information, versus Transformers' \textit{Integrate Late} strategy, which co-locates phoneme, accent, and duration cues in deeper layers (Figure \ref{fig:tsne}).
    \item We show that a logistic regression classifier predicts architecture from peak position profiles alone (AUC = 0.88), demonstrating that representational profiles are architecturally distinctive.
    \item Code and data will be released upon publication to enable reproducibility and extension of our architectural fingerprinting framework.
\end{enumerate}

\section{Methods}
Using linear probing (\S\ref{subsec:tax}), we analyze 24 Transformer and Conformer models (\S\ref{subsec:models}) across over 50,000 utterances from seven speech corpora (\S\ref{subsec:data}). We extract layer-wise representations using standardized depth indexing (\S\ref{subsec:layer_stand}), quantify information accessibility with complementary metrics (\S\ref{subsec:metrics}), and apply statistical tests (\S\ref{subsec:stats}) to isolate architectural effects while controlling for model scale. This approach reveals systematic differences in how architectures organize acoustic, demographic, phonetic, and temporal information.

\subsection{Model Suite}
\label{subsec:models}
We analyzed a diverse collection of 24 pre-trained spoken language models spanning two dominant speech encoder architectural families: Transformers (N=17) and Conformers (N=7). This imbalance reflects the current landscape of publicly available pretrained models and we address its statistical implications in \S\ref{subsec:stats}. The models ranged in size from approximately 39M to 3.3B parameters, providing broad coverage across scales and training regimes. A complete list of all models is provided in Appendix \ref{sec:appendix_details}. All models were analyzed in their publicly released, pre-trained state without fine-tuning.

\subsection{Dataset and Speech Corpora}
\label{subsec:data}
For robustness, we probed model representations using seven diverse speech corpora spanning native and non-native English speakers, multiple accent backgrounds, and varied recording conditions:
\textbf{L2-ARCTIC} (non-native English; \citealp{zhao2018l2arctic}); \textbf{CMU ARCTIC} (native English; \citealp{kominek2004cmu}); \textbf{Common Voice} (accent-stratified crowdsourced; \citealp{ardila2020common}); \textbf{Speech Accent Archive (SAA)} (elicited passage, 100+ L1s; \citealp{weinberger2015speech}); \textbf{ALLSSTAR} (L2 acquisition research; \citealp{bradlow_nd_allsstar}); \textbf{Cambridge Assessment} (private proficiency assessment speech corpus); and \textbf{Speak \& Improve (S\&I) Corpus 2025} (L2 learner English; \citealp{knill2025speak}).

Our corpus comprises 50,000+ utterances across all datasets. Speech samples were processed at each model's native sampling rate (typically 16 kHz) and segmented to match the temporal resolution required for phoneme-level analysis. Results were aggregated across datasets to ensure findings generalize beyond any single corpus. Detailed per-dataset statistics (size, sampling rate, segmentation) are provided in Table \ref{tab:datasets}.

\subsection{Feature Taxonomy and Probing}
\label{subsec:tax}
All analyses in this paper measure \textit{linear accessibility} of information across network depth. Linear probes quantify when a feature can be extracted by a simple readout without additional nonlinear computation. This choice is intentional: linear accessibility reflects functional availability to downstream components and is the standard lens through which representational hierarchies are compared across architectures. We do not claim that linear probes exhaustively measure all information present in a representation; rather, our claims concern when information becomes linearly usable as depth increases.

Following the signal-to-symbol hierarchy established in speech perception research \cite{hickok2007cortical, poeppel2008speech} and validated in neural probing studies \cite{belinkov2017analyzing, pasad2021layer}, we organized features into four levels of abstraction: \textbf{Acoustic} (F0, formants, intensity), \textbf{Demographic} (gender, L1 accent), \textbf{Phonetic} (phoneme category), and \textbf{Temporal} (utterance-level duration). We probe phoneme categories as defined by the CMU pronouncing dictionary (39 categories), which groups allophones of the same phoneme; our probes thus measure categorical sound identity rather than fine-grained allophonic variation. Detailed definitions for each feature are provided in Appendix \ref{sec:appendix_features}. For each model and layer, we trained linear probes to predict these features from hidden states (linear regression for continuous targets; logistic regression for categorical ones). Accordingly, probe curves should be interpreted as measuring \textit{linear accessibility} rather than raw information availability.

\subsection{Depth Standardization}
\label{subsec:layer_stand}
To standardize depth across heterogeneous ASR frontends, we extract the sequence of hidden states output by each encoder block. This yields $L+1$ representations (an initial pre-block representation plus one per encoder block). Throughout, \textbf{Layer 0} denotes the \textit{pre-block} encoder representation (i.e., the embedding/projection output immediately before the first encoder block), and Layer $L$ denotes the final encoder block output. Importantly, this convention \textbf{does not treat convolutional feature-extractor frontends as additional layers}: when present, the feature encoder is upstream of the probed hidden-state stack and is not separately indexed. This ensures comparisons focus on the encoder depth where architectural differences (e.g., per-block convolution vs.\ attention-only blocks) are expressed.
Implementation details for extracting these states across libraries are provided in Appendix \ref{sec:appendix_implementation}.

\subsection{Evaluation Metrics}
\label{subsec:metrics}
We evaluated probing performance using three complementary metrics that capture where in the network a feature is most linearly accessible, how strongly it is encoded at that point, and how separated two features are in depth.

For each feature $f$, we define \textbf{peak position} $\pi_f$ as the normalized depth where probe performance is maximized:
$$
\pi_f = \frac{\arg\max_{l} \text{score}_f(l)}{L}
$$
where $\text{score}_f(l)$ is the test-set accuracy (for classification) or $R^2$ (for regression) at layer $l$, and $L$ is the total number of layers. Peak position ranges from 0 (first layer) to 1 (final layer).

\textbf{Peak strength} $s_f$ is defined as:
$$
s_f = \max_{l} \text{score}_f(l)
$$
which measures the maximum probing performance achieved across layers.

To quantify hierarchical separation between feature pairs, we compute \textbf{positional deltas}:
$$
\Delta_{f_1 \to f_2} = \pi_{f_2} - \pi_{f_1}
$$
Positive deltas indicate that feature $f_2$ peaks later than $f_1$ in the network.

Additional auxiliary metrics, including peak width and layer-wise entropy, are defined in Appendix \ref{sec:appendix_metrics}.

\subsection{Statistical Analysis}
\label{subsec:stats}
\textbf{Architectural Comparison.}
To test whether peak positions differ systematically between architectures, we conducted two-sample t-tests comparing Transformer (N=17) and Conformer (N=7) models for each feature type. We report bootstrap 95\% confidence intervals (10,000 resamples) for the mean difference in peak position.

\textbf{Regression Analysis.}
To isolate the effect of architecture while controlling for model size, we fit a linear regression model:
$\pi_f = \beta_0 + \beta_1 \cdot \text{arch} + \beta_2 \cdot \log(\text{params}) + \epsilon$,
where $\text{arch}$ is a binary indicator (Conformer = 1, Transformer = 0) and $\text{params}$ is the parameter count. We report standardized coefficients ($\beta$) and $p$-values from two-tailed tests.

\textbf{Architectural Fingerprinting.}
To quantify the distinctiveness of architectural fingerprints, we trained a logistic regression classifier to predict architecture (Transformer vs. Conformer) from the 5-dimensional feature vector $[\pi_{\text{acoustic}}, \pi_{\text{gender}}, \pi_{\text{accent}}, \pi_{\text{phoneme}}, \pi_{\text{duration}}]$. We used leave-one-out cross-validation and report the area under the ROC curve (AUC) as our primary metric.

\section{Results}
Our analysis reveals \textbf{systematic, architecture-dependent processing strategies that persist across scale and training regime} in how Transformer and Conformer models organize speech information.

\subsection{Divergent Processing Hierarchies}
Transformer and Conformer models follow distinct orderings of feature representations.
\textbf{Transformers implement an integrated processing strategy} (Acoustic $\rightarrow$ Gender $\rightarrow$ \{Phoneme $\approx$ Duration $\approx$ Accent\}), where high-level linguistic and sociolinguistic information co-locate in layers at 49-57\% of total depth.

\begin{figure}[t]
    \centering
    \includegraphics[width=\columnwidth]{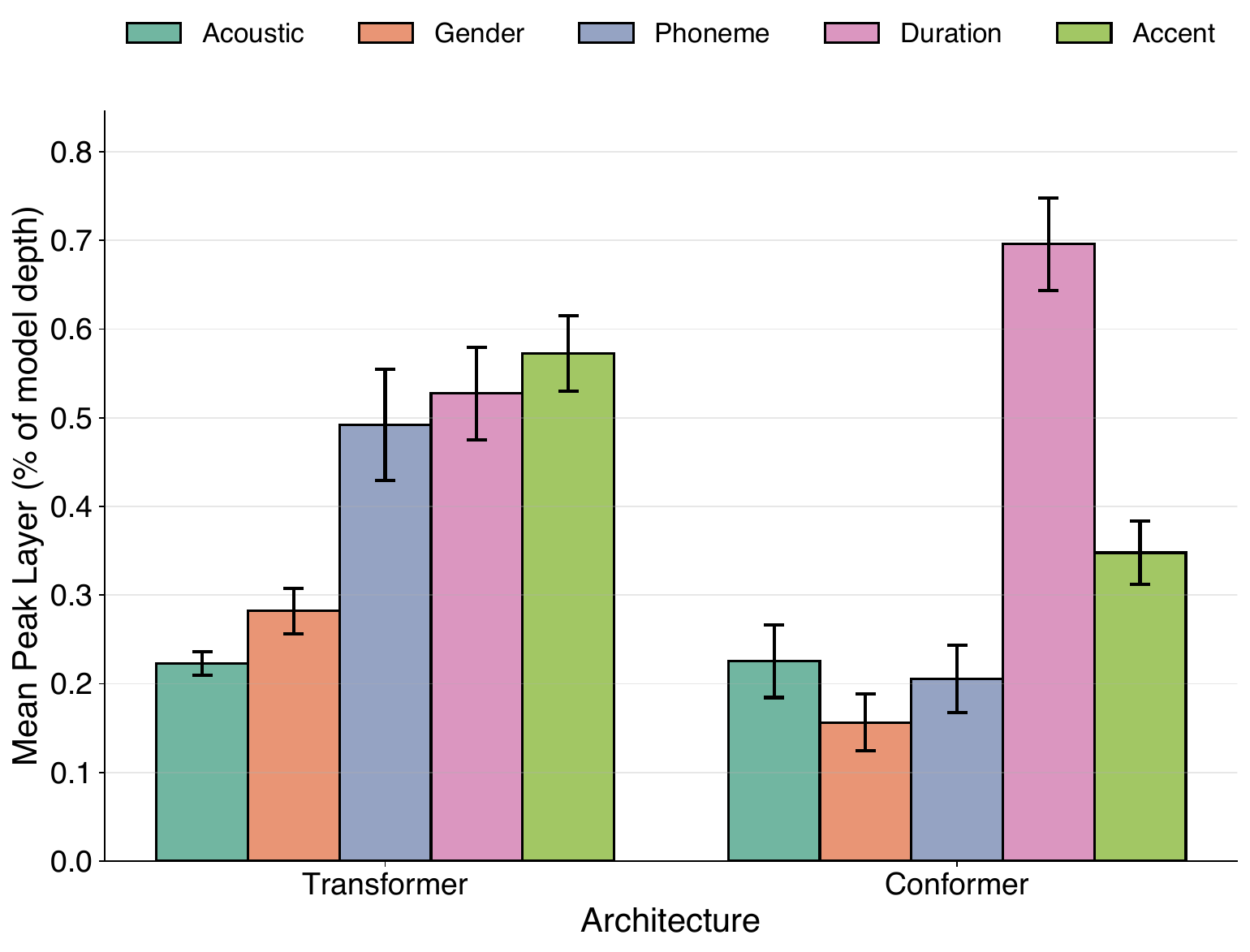}
    \caption{Mean peak layer position (as percentage of model depth) for five feature categories across Transformer (N=17) and Conformer (N=7) architectures. Error bars indicate standard error of the mean. The architectural divergence is clear: Conformers front-load Gender and Phoneme (both $<$ 25\% depth) while deferring Duration to final layers ($\sim$70\% depth), creating a distinct hierarchy. Transformers compress all high-level features into a narrow band (49--57\% depth), reflecting their integrated processing strategy.}
    \label{fig:peak_positions}
\end{figure}

In contrast, \textbf{Conformers exhibit a segregated hierarchy} (Gender $\rightarrow$ Phoneme $\rightarrow$ Acoustic $\rightarrow$ Accent $\rightarrow$ Duration). Notably, Conformers exhibit representational profiles consistent with front-loading categorical features like gender (mean $\pi=0.16$, i.e., 16\% of network depth) and phoneme identity (mean $\pi=0.21$) before resolving fine-grained acoustic details, while deferring temporal integration (duration) to the final layers.
These hierarchies are consistent in our sample: 71\% of Transformers (12/17) show co-located deeper peaks for Phoneme, Duration, and Accent (all $> 35\%$ of normalized depth), while 86\% of Conformers (6/7) show earlier Gender and Phoneme peaks (both $< 50\%$ depth) together with deeper Duration peaks ($> 50\%$ depth). These thresholds are descriptive summaries of the observed distributions in our model set, not theoretical cutoffs. Exceptions are typically driven by unique training regimes (e.g., multilingual multitask training). Figure \ref{fig:peak_positions} summarizes these peak positions across all features, and Table \ref{tab:per_model_positions} provides per-model details showing how individual models within each architectural family conform to these patterns.

\begin{table*}[t]
\centering
\small
\begin{tabular}{lCCCCC}
\toprule
\textbf{Model} & \multicolumn{1}{r}{\textbf{Acoustic}} & \multicolumn{1}{r}{\textbf{Gender}} & \multicolumn{1}{r}{\textbf{Accent}} & \multicolumn{1}{r}{\textbf{Phoneme}} & \multicolumn{1}{r}{\textbf{Duration}} \\
\midrule
\multicolumn{6}{c}{\textit{Conformer Models}} \\
\midrule
canary-1b & 0.153 & 0.139 & 0.347 & 0.215 & 0.611 \\
canary-1b-flash & 0.146 & 0.089 & 0.260 & 0.130 & 0.849 \\
canary-qwen-2.5b & 0.151 & 0.099 & 0.255 & 0.312 & 0.464 \\
granite-speech-3.3-2b & 0.466 & 0.323 & 0.375 & 0.229 & 0.812 \\
parakeet-tdt-0.6b-v2 & 0.194 & 0.167 & 0.340 & 0.340 & 0.611 \\
speechbrain-loq & 0.206 & 0.083 & 0.537 & 0.157 & 0.759 \\
w2v2-conformer & 0.192 & 0.194 & 0.319 & 0.056 & 0.764 \\
\midrule
\multicolumn{6}{c}{\textit{Transformer Models}} \\
\midrule
Phi-4-multimodal & 0.332 & 0.281 & 0.266 & 0.896 & 0.109 \\
hubert-large & 0.140 & 0.181 & 0.278 & 0.000 & 0.681 \\
hubert-xlarge & 0.091 & 0.073 & 0.271 & 0.156 & 0.538 \\
wav2vec2-large & 0.137 & 0.090 & 0.278 & 0.014 & 0.424 \\
wavlm-large & 0.186 & 0.118 & 0.618 & 0.500 & 0.667 \\
whisper-base & 0.306 & 0.333 & 0.722 & 0.472 & 0.472 \\
whisper-base.en & 0.258 & 0.278 & 0.667 & 0.778 & 0.472 \\
whisper-large & 0.217 & 0.297 & 0.698 & 0.245 & 0.552 \\
whisper-large-v2 & 0.202 & 0.328 & 0.667 & 0.797 & 0.990 \\
whisper-large-v3 & 0.202 & 0.406 & 0.708 & 0.604 & 1.000 \\
whisper-large-v3-turbo & 0.208 & 0.396 & 0.651 & 0.495 & 0.380 \\
whisper-medium & 0.172 & 0.292 & 0.618 & 0.417 & 0.465 \\
whisper-medium.en & 0.238 & 0.403 & 0.750 & 0.611 & 0.493 \\
whisper-small & 0.225 & 0.306 & 0.639 & 0.542 & 0.403 \\
whisper-small.en & 0.214 & 0.306 & 0.653 & 0.625 & 0.486 \\
whisper-tiny & 0.208 & 0.375 & 0.667 & 0.667 & 0.458 \\
whisper-tiny.en & 0.262 & 0.333 & 0.583 & 0.542 & 0.375 \\
\bottomrule
\end{tabular}
\caption{Mean peak layer position for each model. Values represent the fraction of model depth (0 = first layer, 1 = final layer) where probing accuracy/R$^2$ is maximized. Conformers consistently show earlier Gender and Phoneme peaks compared to Transformers, while Duration peaks later in Conformers, reflecting their staged ``Categorize Early'' processing strategy.}
\label{tab:per_model_positions}
\end{table*}

\subsection{Positional Dynamics of Individual Features}
We examine how architectural inductive biases shape the encoding of specific features. These feature categories are not independent: speech cues often covary across acoustic, demographic, phonemic, and prosodic dimensions. This covariance motivated our focus on depth-wise separability, since the goal is not to assume isolated signals, but to test whether architectures differ in when particular information types become linearly accessible. Despite this overlap, we still observe clear differences in relative positioning; for example, in Conformers, Gender peaks earlier (mean $\pi=0.16$) than acoustic features such as F0 (mean $\pi=0.23$).

\textbf{Acoustic features} are front-loaded in both architectures (Transformer mean $\pi=0.22$; Conformer mean $\pi=0.23$), as expected for low-level signals.

\textbf{Gender} peaks very early in Conformers (mean $\pi=0.16$), often \textit{before} fine-grained acoustic features ($\pi=0.23$), while Transformers distribute gender encoding more broadly (mean $\pi=0.28$). Note that our use of binary gender labels reflects dataset annotations and may not capture the full spectrum of gender identity; similar caveats apply to coarse accent categories.

\textbf{Phoneme category} shows the largest divergence ($\Delta \pi = -0.29$, $p < 0.01$). Conformers peak early (mean $\pi=0.21$) while Transformers peak late (mean $\pi=0.49$).

\textbf{Accent} peaks earlier in Conformers (mean $\pi=0.35$) than Transformers (mean $\pi=0.57$).

\textbf{Duration}, requiring the longest-range integration, represents the final stage of processing for both architectures (Transformer $\pi=0.53$; Conformer $\pi=0.70$). We also found a universal hierarchy in peak probing performance (Gender > Duration > Accent > Acoustic > Phoneme) across architectures; detailed results on representational entropy and outliers are provided in Appendix \ref{sec:appendix_additional_results}.

Figure \ref{fig:layerwise_trajectories} visualizes these dynamics as smoothed probing trajectories across normalized depth. The Conformer curves show sharp early rises for Gender and Phoneme followed by plateau, while Transformer curves rise more gradually and peak in mid-to-deep layers. These trajectories underlie the summary statistics in Figure \ref{fig:peak_positions} and Table \ref{tab:per_model_positions}.

\begin{figure*}[t]
    \centering
    \includegraphics[width=\textwidth]{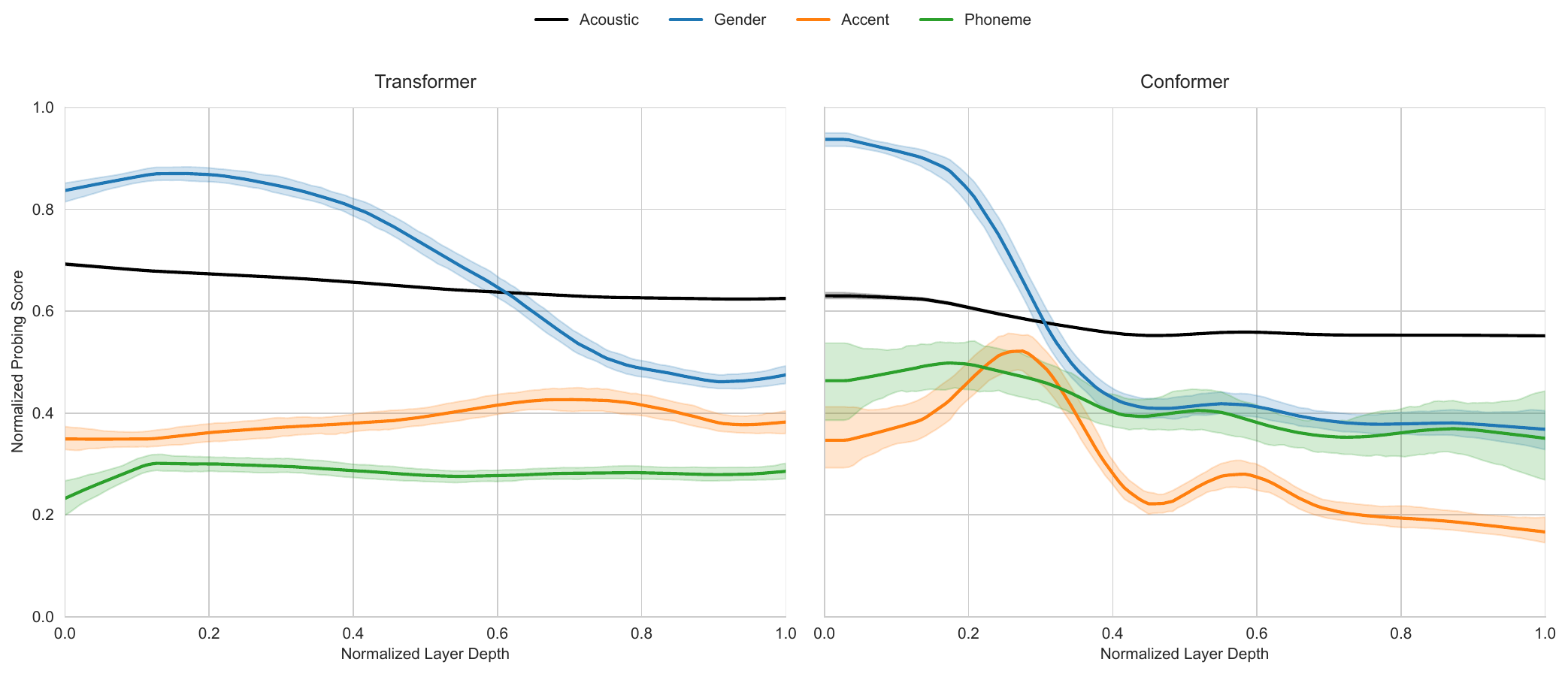}
    \caption{Smoothed layer-wise probing trajectories across normalized depth for representative feature groups (Acoustic, Gender, Accent, Phoneme), shown separately for Transformers (left) and Conformers (right). Curves are LOWESS fits with 95\% bootstrapped confidence intervals. To emphasize \textit{where} features become accessible (rather than absolute probe difficulty), probing scores within each feature group are min--max normalized before smoothing. Duration is omitted from this visualization for clarity; its peak positions are reported in Figure \ref{fig:peak_positions} and the main text.}
    \label{fig:layerwise_trajectories}
\end{figure*}

\subsection{Architectural Fingerprints}
The systematic differences we observe allow us to classify a model's architecture from its representational profile alone with high accuracy (AUC = 0.88). The most discriminative features are accent ($\beta = -0.85$), gender ($\beta = -0.83$), and duration ($\beta = +0.81$). Controlling for model scale confirms these effects are architectural: \textbf{accent shows a significant architecture coefficient} ($\beta_{\text{arch}} = -0.21$, $p < 0.05$) independent of parameter count. Table \ref{tab:ttests} reports the complete statistical comparison, showing large effect sizes (Cohen's $d > 0.8$) for all features except Acoustic. With Bonferroni correction ($\alpha = 0.01$), Accent and Phoneme differences remain significant. Full statistical details including bootstrap confidence intervals are in Appendix \ref{sec:appendix_stats_detailed}.

\begin{table}[t]
\centering
\small
\begin{tabular}{lrrrrr}
\toprule
\textbf{Feature} & \textbf{$\Delta \pi$} & \textbf{$t$} & \textbf{df} & \textbf{$p$} & \textbf{Cohen's $d$} \\
\midrule
Acoustic & +0.003 & $-$0.08 & 22 & 0.940 & 0.03 \\
Gender & $-$0.126 & 2.80 & 22 & 0.011 & $-$1.26 \\
Accent & $-$0.225 & 3.17 & 22 & 0.004 & $-$1.42 \\
Phoneme & $-$0.286 & 2.81 & 22 & 0.010 & $-$1.26 \\
Duration & +0.168 & $-$1.90 & 22 & 0.071 & 0.85 \\
\bottomrule
\end{tabular}
\caption{T-tests comparing peak positions between architectures. $\Delta \pi$ = Conformer $-$ Transformer (negative values indicate Conformers peak earlier). Cohen's $d > 0.8$ indicates large effect size. With Bonferroni correction for 5 comparisons ($\alpha = 0.01$), Accent and Phoneme survive correction.}
\label{tab:ttests}
\end{table}

\subsection{Architecture vs. Training Regime}
While architecture explains most variance, training regime modulates specific features, as evident in the per-model peak positions shown in Table \ref{tab:per_model_positions}.

Rather than treating Whisper as an anomaly, we interpret it as evidence for a \textit{hierarchy of controls}: \textbf{architecture sets the default representational profile}, while \textbf{objective acts as a modulator}. Whisper models (supervised, multilingual) delay gender and accent processing significantly compared to SSL Transformers (HuBERT/WavLM), consistent with a multitask objective that benefits from retaining indexical cues deeper in the network. However, the architectural fingerprint persists: non-Whisper Transformers still exhibit the ``Integrate Late'' strategy relative to Conformers (details in Appendix \ref{sec:appendix_outliers}).

\subsection{Robustness and Controls}
To ensure our findings are not driven by outliers or specific training factors, we conducted extensive robustness checks (detailed in Appendix \ref{sec:appendix_robustness}). A leave-one-out sensitivity analysis confirms that architectural effects become \textit{stronger} when excluding the outlier model \textit{granite-speech-3.3-2b} (e.g., gender $\Delta \pi$ improves from -0.13 to -0.15, $p=0.003$). Additionally, a paired comparison of multilingual vs. English-only Whisper models reveals no significant difference in peak positions for any feature (though phoneme shows a non-significant trend, $\Delta\pi = -0.12$, $p = 0.30$), indicating that the delayed processing in Whisper models is likely due to their multitask objective rather than multilingual data diversity.

\section{Discussion}
Our findings demonstrate that high-performance speech recognition emerges from fundamentally different processing strategies shaped by architectural inductive bias, not convergence to a single solution. The consistency of these architectural fingerprints, observable across model scales, training regimes, and even within sub-families, suggests we have uncovered organizing principles rather than superficial correlations. The ability to classify architecture from representational profiles alone (AUC = 0.88) underscores how inductive bias shapes the learned solution.

Prior probing work has established that self-supervised speech models encode features hierarchically \cite{pasad2021layer, pasad_comparative_2023}. Early work demonstrated that neural models of speech develop linguistically meaningful representations across layers, providing an initial interpretability perspective on spoken representations. However, these studies primarily analyzed models individually without systematic architectural comparison. Our architecture-controlled analysis reveals that similar task performance can arise from fundamentally different processing strategies.

Work in other domains has shown that architectures with similar performance can nevertheless differ substantially in their internal representations. For example, \citet{raghu2021dovisiontransformers} compare Vision Transformers and CNNs and find clear differences in layer-wise representation structure. Our results extend this comparative perspective to speech, showing that Transformer and Conformer encoders exhibit distinct representational timelines despite comparable ASR performance.

Taken together, our findings demonstrate that benchmark parity does not imply representational convergence---a result with implications for interpretability, transfer learning, and model selection.

\subsection*{Why Do These Divergent Strategies Emerge?}
We propose that the observed differences arise from a distinction in how each architecture processes local versus global information.

\xhdr{Conformers' ``Categorize Early'' Strategy} The Conformer's depthwise separable convolutions may provide a strong inductive bias for local pattern detection. This is consistent with the rapid linear accessibility of categorical features (gender from coarse spectral patterns, phonemes from local acoustic templates) we observe in early layers. The staged pipeline, with 54\% depth separation between gender and duration, suggests later layers perform higher-order integration.

\xhdr{Transformers' ``Integrate Late'' Strategy} Pure self-attention lacks explicit local bias, which may require the network to learn local patterns implicitly through attention weights. This is consistent with the more gradual, distributed processing flow we observe. The co-location of accent, phoneme, and duration in the latter half of the network (all peaking between 49-57\% depth) suggests these models resolve multiple feature types concurrently rather than through staged processing.

\xhdr{Objectives Modulate Architecture} The Whisper Effect (§3.4) demonstrates that training objectives can shift representational timing within an architectural family. This supports a ``hierarchy of controls'': \textbf{architecture sets the default processing strategy; training objectives modulate it}. For practitioners, this suggests that architectural choice determines the \textit{shape} of the representational profile, while objective choice fine-tunes \textit{where} specific features peak. We emphasize that these are testable hypotheses grounded in linear accessibility patterns, not causal claims about mechanism.

\subsection*{Implications for Model Design}
Our findings suggest concrete selection heuristics:
\begin{itemize}
    \item \textbf{Streaming and low-latency}: Conformers' early phoneme category accessibility (21\% depth vs.\ 49\% for Transformers) means useful representations are available sooner. We predict Conformer-based systems will show smaller WER degradation under aggressive layer truncation or streaming constraints.
    \item \textbf{Speaker variability}: Transformers co-locate gender and accent with phoneme categories in deep layers, suggesting tighter integration of speaker normalization. We predict Transformer-based systems will show smaller performance gaps across speaker demographics when using full-depth representations. Recent work by \citet{roll2025scaling} supports this hypothesis, finding that while Conformer-based models achieve lower overall WER, they exhibit significantly higher relative error rates on non-native speech (``conformation bias'') compared to Transformers.
    \item \textbf{Layer-wise distillation}: The staged Conformer hierarchy (gender $\rightarrow$ phoneme $\rightarrow$ acoustic $\rightarrow$ accent $\rightarrow$ duration) suggests natural cut-points for layer-wise knowledge distillation; the integrated Transformer profile may require different distillation strategies.
\end{itemize}

\xhdr{Future Directions}
Hybrid architectures combining Conformer-style early convolutions with Transformer-style deep integration capture benefits of both strategies, while training objectives may enable control over when features become accessible.

\section{Conclusion}
We demonstrated that state-of-the-art spoken language models do not converge on a single processing strategy; their \textbf{internal hierarchies are shaped by architectural inductive bias}. We identified two divergent pathways: a \textit{segregated} ``categorize early'' hierarchy in Conformers, where gender and phoneme categories become linearly accessible in shallow layers before higher-order integration, and an \textit{integrated, distributed} ``integrate late'' hierarchy in Transformers, which co-locates phoneme, accent, and duration cues in deeper layers.

These differences are quantifiable: phoneme category reaches its peak 29\% earlier in Conformers than Transformers, representing the largest architectural effect we observed and surviving Bonferroni correction. In contrast, duration exhibits a trend toward later processing, peaking 17\% deeper ($p = 0.071$). We show that a simple classifier can distinguish model families from representational profiles alone (AUC = 0.88, 95\% CI [0.73, 1.00]), reinforcing the existence of stable architectural fingerprints (though the wide confidence interval reflects our moderate sample size).

Our findings reveal that high-performance speech recognition can emerge from different computational solutions. The path to expert performance is not singular: Conformers' local-then-global inductive bias and Transformers' global-only attention lead to different but equally effective processing strategies. This has practical implications for model selection (Conformers may excel at speaker-invariant tasks; Transformers at speaker-dependent ones) and underscores the impact of architectural priors on learned representations, paving the way for more intentional model design in speech processing. We envision Architectural Fingerprinting as a diagnostic framework for model developers: as speech models continue to scale, it offers a way to look beyond standard performance metrics (e.g., WER) and assess whether a model's representations and processing strategy align with its intended use.

\section*{Limitations}

\textbf{Linearity vs.\ availability.} Linear probes measure \textit{linear accessibility} and cannot distinguish late \textit{integration} from late \textit{linearization}. Our claims are thus restricted to when information becomes linearly usable, not when it first emerges in any form. Testing nonlinear accessibility (as proposed in Falsifiable Predictions) would refine the interpretation of late accessibility without altering the architectural distinctions reported here.

\textbf{Error pattern analysis.} Our analysis characterizes \textit{when} features emerge but not \textit{how} they are represented. We cannot distinguish, for example, whether early Conformer phoneme encoding reflects local template matching or a different mechanism. Analyzing confusion matrices of phoneme probes could address this limitation, but such mechanistic validation is beyond our current scope.

\textbf{Feature taxonomy scope.} Our five-feature taxonomy captures the signal-to-phoneme transformation central to ASR but does not address lexical, syntactic, or discourse-level organization. Extending this to word-level semantics would provide a more complete picture of speech understanding.

\textbf{Native vs.\ non-native speech.} Our corpus includes both native speakers (CMU ARCTIC) and non-native speakers (L2-ARCTIC and others), but we do not systematically test for interactions between architecture and speaker nativeness. This limits our ability to characterize how architectural differences affect robustness to pronunciation variation, despite evidence that human listeners show processing delays for non-native accents \cite{munro_processing_1995}.

\textbf{Model diversity and sample imbalance.} Our model suite is unbalanced across architectures, with fewer Conformers (N=7) than Transformers (N=17), largely due to the present distribution of publicly available pretrained ASR models. This limits the strength of claims about within-architecture variability, especially for Conformers, and may reduce sensitivity to less common representational profiles. We therefore interpret our conclusions at the level of broad architectural tendencies rather than exhaustive coverage of each family, and we include robustness checks in Appendix D showing that the main Transformer--Conformer differences remain stable under leave-one-out analyses. Our suite also does not include transducers \cite{graves2012sequence} or state-space models, limiting generalization to the full ASR landscape.

\textbf{Correlational nature.} Our analysis is correlational: architecture predicts hierarchy, but cannot definitively establish causation. Controlled experiments (e.g., adding convolutions to Transformers) would provide stronger evidence but require training models from scratch, a substantial undertaking given the scale of modern ASR systems.

\textbf{Ethical considerations.} Our use of binary gender labels and coarse accent categories reflects dataset annotations but does not capture the full spectrum of gender identity or accent diversity. Probing can expose sensitive attributes from representations, raising privacy concerns for deployment. We encourage practitioners to conduct fairness audits before using fingerprinting for model selection.

\bibliography{custom}

\appendix

\section{Experimental Details}
\label{sec:appendix_details}

\subsection{Model Suite}
Our analysis included 24 pre-trained models from the Transformer and Conformer families. Table \ref{tab:model_suite_detailed} provides comprehensive architectural specifications.

\begin{table*}[t]
\centering
\small
\begin{tabular}{llrrrrp{3.5cm}}
\toprule
\textbf{Model} & \textbf{Arch} & \textbf{Params} & \textbf{Layers} & \textbf{Hidden} & \textbf{Heads} & \textbf{Training} \\
\midrule
\multicolumn{7}{c}{\textit{Transformer Models (N=17)}} \\
\midrule
whisper-tiny & Trans. & 39M & 4 & 384 & 6 & Supervised, 680k hrs \\
whisper-tiny.en & Trans. & 39M & 4 & 384 & 6 & Supervised, English-only \\
whisper-base & Trans. & 74M & 6 & 512 & 8 & Supervised, 680k hrs \\
whisper-base.en & Trans. & 74M & 6 & 512 & 8 & Supervised, English-only \\
whisper-small & Trans. & 244M & 12 & 768 & 12 & Supervised, 680k hrs \\
whisper-small.en & Trans. & 244M & 12 & 768 & 12 & Supervised, English-only \\
whisper-medium & Trans. & 769M & 24 & 1024 & 16 & Supervised, 680k hrs \\
whisper-medium.en & Trans. & 769M & 24 & 1024 & 16 & Supervised, English-only \\
whisper-large & Trans. & 1.55B & 32 & 1280 & 20 & Supervised, 680k hrs \\
whisper-large-v2 & Trans. & 1.55B & 32 & 1280 & 20 & Supervised, 680k hrs \\
whisper-large-v3 & Trans. & 1.55B & 32 & 1280 & 20 & Supervised, 1M+ hrs \\
whisper-large-v3-turbo & Trans. & 809M & 32 & 1280 & 20 & Supervised, distilled \\
wav2vec2-large-960h-lv60 & Trans. & 317M & 24 & 1024 & 16 & SSL + fine-tuned \\
hubert-large-ls960-ft & Trans. & 317M & 24 & 1024 & 16 & SSL + fine-tuned \\
hubert-xlarge-ls960-ft & Trans. & 1B & 48 & 1280 & 16 & SSL + fine-tuned \\
wavlm-large & Trans. & 317M & 24 & 1024 & 16 & SSL \\
Phi-4-multimodal-instruct & Trans. & 2.7B & 40 & 3072 & 32 & Supervised, multimodal \\
\midrule
\multicolumn{7}{c}{\textit{Conformer Models (N=7)}} \\
\midrule
wav2vec2-conformer & Conf. & 430M & 24 & 1024 & 16 & SSL \\
canary-1b & Conf. & 1B & 24 & 1024 & 8 & Supervised, multilingual \\
canary-1b-flash & Conf. & 1B & 24 & 1024 & 8 & Supervised, distilled \\
canary-qwen-2.5b & Conf. & 2.5B & 40 & 1536 & 12 & Supervised, Qwen decoder \\
parakeet-tdt-0.6b-v2 & Conf. & 600M & 30 & 768 & 8 & Supervised, TDT \\
granite-speech-3.3-2b & Conf. & 3.3B & 32 & 2048 & 16 & Supervised, code-switched \\
speechbrain-loq & Conf. & 135M & 12 & 512 & 8 & Supervised \\
\bottomrule
\end{tabular}
\caption{Comprehensive model specifications. Trans. = Transformer, Conf. = Conformer. SSL = self-supervised learning. All Whisper models from OpenAI, HuBERT/Wav2Vec2 from Meta, WavLM from Microsoft, Canary/Parakeet from NVIDIA, Granite from IBM, SpeechBrain from community.}
\label{tab:model_suite_detailed}
\end{table*}

\subsection{Dataset Details}
We aggregated probing results across seven speech corpora to ensure generalizability. Table \ref{tab:datasets} summarizes each dataset.

\begin{table}[h]
\centering
\small
\resizebox{\columnwidth}{!}{
\begin{tabular}{lrrll}
\toprule
\textbf{Dataset} & \textbf{Utterances} & \textbf{Rate (kHz)} & \textbf{Description} & \textbf{Segmentation} \\
\midrule
L2-ARCTIC & 3,750 & 44.1 & Non-native English, 6 L1s & MFA forced-aligned \\
CMU ARCTIC & 4,500 & 16 & Native English, regional accents & MFA forced-aligned \\
Common Voice & 10,000+ & 48 & Crowdsourced, accent-stratified & MFA forced-aligned \\
SAA & 2,000+ & 44.1 & Elicited passage, 100+ L1s & MFA forced-aligned \\
ALLSSTAR & 5,000+ & 44.1 & L2 acquisition research & MFA forced-aligned \\
Cambridge & 15,000+ & 16 & Proficiency assessment & MFA forced-aligned \\
S\&I Corpus 2025 & 3,000+ & 16 & L2 learner English, proficiency assessment & MFA forced-aligned \\
\midrule
\textbf{Total} & \textbf{$\sim$50,000} & -- & & \\
\bottomrule
\end{tabular}
}
\caption{Speech corpora used for probing. Utterance counts are approximate. All audio was resampled to 16 kHz for model inference. MFA = Montreal Forced Aligner.}
\label{tab:datasets}
\end{table}

Results were computed per-dataset and then aggregated (mean peak positions across datasets) to ensure findings are not driven by idiosyncrasies of any single corpus. The L2-ARCTIC dataset \cite{zhao2018l2arctic} was used for t-SNE visualizations (Figure \ref{fig:tsne}) due to its balanced L1 representation.

\subsection{Feature Extraction Details}
\subsubsection{Acoustic Features (24 probes)}
Acoustic features were extracted using Praat \cite{boersma2001praat} with a 10ms frame shift:
\begin{itemize}
    \item \textbf{F0} (fundamental frequency): min, mean, median, max
    \item \textbf{F1, F2, F3} (formants): min, mean, median, max each
    \item \textbf{F3$-$F2} (formant dispersion): min, mean, median, max
    \item \textbf{Intensity}: min, mean, median, max
\end{itemize}

\subsubsection{Demographic Features (2 probes)}
\begin{itemize}
    \item \textbf{Gender}: Binary classification (male/female).
    \item \textbf{L1 Background}: 6-way classification (Arabic, Hindi, Korean, Mandarin, Spanish, Vietnamese). We standardized L1 labels to these categories based on speaker metadata. For corpora with self-reported accent labels (e.g., Common Voice), we mapped to the nearest L1 category. Corpora without L1 metadata were excluded from accent probes.
\end{itemize}
For utterance-level labels (gender, accent), we applied the label to all frames within the utterance. Probes were trained at the frame level, but test accuracy was computed per-frame and then averaged per-utterance to avoid inflating sample sizes. We did not control for differences in frame stride across models, which is a limitation.

\subsubsection{Linguistic Features (2 probes)}
\begin{itemize}
    \item \textbf{Phoneme category}: 39-way classification (CMU phoneme set, which groups allophones). Phoneme boundaries were obtained using the Montreal Forced Aligner (MFA) with the English pronunciation dictionary. For corpora lacking transcripts, we used Whisper-large-v3 for automatic transcription prior to alignment.
    \item \textbf{Duration}: Regression (utterance-level duration in ms)
\end{itemize}

\subsection{Linear Probe Training}
Linear probes were trained independently for each feature at each layer:
\begin{itemize}
    \item \textbf{Classification tasks}: Cross-entropy loss (Gender, L1, Phoneme)
    \item \textbf{Regression tasks}: Mean squared error loss (Acoustic, Duration)
    \item \textbf{Optimizer}: Adam with learning rate 0.001
    \item \textbf{Batch size}: 32
    \item \textbf{Max epochs}: 50 with early stopping
    \item \textbf{Data split}: 80\%/10\%/10\% train/validation/test. For demographic probes (gender, accent), we used speaker-disjoint splits where possible (L2-ARCTIC, CMU ARCTIC). For corpora without speaker metadata, we applied random splits; this may inflate early-layer accuracy for these features, and we note this as a limitation.
\end{itemize}

\subsection{Implementation Details}
\label{sec:appendix_implementation}
To extract hidden states consistently across models, we utilized the standard interface provided by the HuggingFace Transformers library. Specifically, models were invoked with \texttt{output\_hidden\_states=True}, returning a tuple containing the output of the embedding layer followed by the output of each Transformer or Conformer block. For models not in HuggingFace (e.g., some SpeechBrain or ESPnet models), we manually instrumented the forward pass to capture the equivalent intermediate representations.

\section{Detailed Statistical Results}
\label{sec:appendix_stats}
\label{sec:appendix_stats_detailed}

\subsection{Per-Model Peak Scores}
Table \ref{tab:per_model_scores} shows the maximum probing score achieved for each model across feature categories.

\begin{table*}[t]
\centering
\small
\begin{tabular}{lCCCCC}
\toprule
\textbf{Model} & \multicolumn{1}{r}{\textbf{Acoustic}} & \multicolumn{1}{r}{\textbf{Gender}} & \multicolumn{1}{r}{\textbf{Accent}} & \multicolumn{1}{r}{\textbf{Phoneme}} & \multicolumn{1}{r}{\textbf{Duration}} \\
\midrule
\multicolumn{6}{c}{\textit{Conformer Models}} \\
\midrule
canary-1b & 0.303 & 0.962 & 0.558 & 0.094 & 0.761 \\
canary-1b-flash & 0.305 & 0.960 & 0.574 & 0.079 & 0.834 \\
canary-qwen-2.5b & 0.302 & 0.959 & 0.557 & 0.107 & 0.748 \\
granite-speech-3.3-2b & 0.066 & 0.568 & 0.201 & 0.074 & 0.873 \\
parakeet-tdt-0.6b-v2 & 0.307 & 0.951 & 0.554 & 0.079 & 0.727 \\
speechbrain-loq & 0.321 & 0.951 & 0.519 & 0.103 & 0.758 \\
w2v2-conformer & 0.311 & 0.962 & 0.461 & 0.168 & 0.786 \\
\midrule
\multicolumn{6}{c}{\textit{Transformer Models}} \\
\midrule
Phi-4-multimodal & 0.063 & 0.559 & 0.257 & 0.059 & 0.813 \\
hubert-large & 0.302 & 0.939 & 0.452 & 0.133 & 0.807 \\
hubert-xlarge & 0.302 & 0.938 & 0.466 & 0.102 & 0.823 \\
wav2vec2-large & 0.285 & 0.880 & 0.442 & 0.147 & 0.806 \\
wavlm-large & 0.276 & 0.883 & 0.514 & 0.118 & 0.785 \\
whisper-base & 0.308 & 0.952 & 0.467 & 0.083 & 0.913 \\
whisper-base.en & 0.301 & 0.917 & 0.464 & 0.082 & 0.901 \\
whisper-large & 0.339 & 0.943 & 0.592 & 0.095 & 0.917 \\
whisper-large-v2 & 0.342 & 0.966 & 0.590 & 0.085 & 0.928 \\
whisper-large-v3 & 0.343 & 0.959 & 0.582 & 0.063 & 0.929 \\
whisper-large-v3-turbo & 0.354 & 0.961 & 0.589 & 0.063 & 0.915 \\
whisper-medium & 0.337 & 0.952 & 0.587 & 0.088 & 0.900 \\
whisper-medium.en & 0.338 & 0.958 & 0.543 & 0.102 & 0.911 \\
whisper-small & 0.331 & 0.959 & 0.494 & 0.096 & 0.911 \\
whisper-small.en & 0.329 & 0.951 & 0.465 & 0.095 & 0.923 \\
whisper-tiny & 0.310 & 0.937 & 0.406 & 0.081 & 0.904 \\
whisper-tiny.en & 0.313 & 0.939 & 0.394 & 0.086 & 0.883 \\
\bottomrule
\end{tabular}
\caption{Peak probing scores for each model. Acoustic/Duration values are R$^2$; Gender/Accent/Phoneme values are accuracy.}
\label{tab:per_model_scores}
\end{table*}

\subsection{Complete Statistical Tests}
\label{sec:appendix_tests}

\subsubsection{Bootstrap Confidence Intervals}
\begin{table}[h]
\centering
\small
\begin{tabular}{lrrr}
\toprule
\textbf{Feature} & \textbf{$\Delta \pi$} & \textbf{95\% CI Low} & \textbf{95\% CI High} \\
\midrule
Acoustic & +0.002 & $-$0.061 & +0.090 \\
Gender & $-$0.126 & $-$0.196 & $-$0.046 \\
Accent & $-$0.224 & $-$0.321 & $-$0.117 \\
Phoneme & $-$0.286 & $-$0.423 & $-$0.146 \\
Duration & +0.168 & +0.029 & +0.298 \\
\bottomrule
\end{tabular}
\caption{Bootstrap 95\% confidence intervals (10,000 resamples) for mean peak position differences between architectures.}
\label{tab:bootstrap}
\end{table}

\subsubsection{Regression Controlling for Model Size}
\begin{table}[h]
\centering
\small
\begin{tabular}{lrrrr}
\toprule
\textbf{Feature} & \textbf{$\beta_{\text{arch}}$} & \textbf{$p_{\text{arch}}$} & \textbf{$\beta_{\text{size}}$} & \textbf{$p_{\text{size}}$} \\
\midrule
Acoustic & +0.016 & 0.729 & +0.004 & 0.803 \\
Gender & $-$0.105 & 0.116 & +0.002 & 0.908 \\
Accent & $-$0.214 & 0.041 & $-$0.024 & 0.425 \\
Phoneme & $-$0.257 & 0.096 & $-$0.006 & 0.901 \\
Duration & +0.146 & 0.240 & +0.042 & 0.276 \\
\bottomrule
\end{tabular}
\caption{Linear regression: $\pi_f = \beta_0 + \beta_{\text{arch}} \cdot \text{Conformer} + \beta_{\text{size}} \cdot \log(\text{params})$.}
\label{tab:regression}
\end{table}

\subsection{Architectural Classifier Details}
The logistic regression classifier for predicting architecture achieved:
\begin{itemize}
    \item \textbf{AUC}: 0.88 (95\% CI [0.73, 1.00])
    \item \textbf{Leave-one-out accuracy}: 19/24 = 79.2\%
\end{itemize}

\noindent Standardized coefficients:
\begin{itemize}
    \item Accent position: $\beta = -0.85$ (most discriminative)
    \item Gender position: $\beta = -0.83$
    \item Duration position: $\beta = +0.81$
    \item Acoustic position: $\beta = +0.68$
    \item Phoneme position: $\beta = -0.33$
\end{itemize}

\subsection{Acoustic Feature Breakdown}
Table \ref{tab:acoustic_breakdown} shows detailed statistics for individual acoustic features.

\begin{table}[h]
\centering
\scriptsize
\begin{tabular}{lrrrr}
\toprule
\textbf{Feature} & \multicolumn{2}{c}{\textbf{Transformer}} & \multicolumn{2}{c}{\textbf{Conformer}} \\
 & \textbf{Pos} & \textbf{Score} & \textbf{Pos} & \textbf{Score} \\
\midrule
F0 mean & 0.167 & 0.662 & 0.067 & 0.651 \\
F0 median & 0.186 & 0.708 & 0.075 & 0.693 \\
F0 min & 0.286 & 0.179 & 0.283 & 0.187 \\
F0 max & 0.248 & 0.113 & 0.267 & 0.098 \\
F1 mean & 0.119 & 0.286 & 0.183 & 0.226 \\
F1 median & 0.138 & 0.283 & 0.200 & 0.239 \\
F2 mean & 0.154 & 0.280 & 0.170 & 0.346 \\
F3 mean & 0.138 & 0.362 & 0.114 & 0.346 \\
F3 median & 0.180 & 0.466 & 0.113 & 0.463 \\
Intensity max & 0.088 & 0.640 & 0.209 & 0.348 \\
Intensity mean & 0.152 & 0.532 & 0.154 & 0.376 \\
\bottomrule
\end{tabular}
\caption{Selected acoustic feature positions and scores by architecture. Pos = mean peak layer position; Score = mean peak R$^2$.}
\label{tab:acoustic_breakdown}
\end{table}

\subsection{Whisper Model Family Analysis}
\begin{table}[H]
\centering
\small
\begin{tabular}{lrrr}
\toprule
\textbf{Feature} & \textbf{Whisper} & \textbf{Non-Whisper} & \textbf{$p$} \\
\midrule
Gender & 0.338 & 0.149 & $<$0.001 \\
Accent & 0.669 & 0.342 & $<$0.001 \\
Duration & 0.546 & 0.484 & 0.606 \\
\bottomrule
\end{tabular}
\caption{Whisper (N=12) vs. non-Whisper Transformer (N=5) peak positions. Whisper models show significantly later gender and accent peaks.}
\label{tab:whisper}
\end{table}

\section{Additional Metric Definitions}
\label{sec:appendix_metrics}

\subsubsection{Peak Width}
To measure whether features are sharply localized or broadly distributed, we computed \textbf{peak width} $w_f$ as the proportion of layers where performance exceeds 70\% of peak strength:
$$
w_f = \frac{1}{L+1} \sum_{l=0}^{L} \mathbb{1}[\text{score}_f(l) \geq 0.7 \cdot s_f]
$$
Low values indicate sharp, localized peaks; high values indicate distributed encoding.

\subsubsection{Layer-wise Entropy}
To measure the concentration of feature information, we computed the entropy of the normalized performance distribution across layers:
$$
H_f = -\sum_{l=0}^{L} q_f(l) \log q_f(l)
$$
where
$$
q_f(l) = \frac{\text{score}_f(l)}{\sum_{l'} \text{score}_f(l')}
$$
is the normalized performance at layer $l$. High entropy indicates distributed encoding; low entropy indicates concentration in specific layers.

\section{Robustness Checks}
\label{sec:appendix_robustness}
\subsection{Sensitivity Analysis: Robustness to Outliers}
Given the small Conformer sample size (N=7) and the previously noted granite-speech-3.3-2b anomaly, we conducted a leave-one-out sensitivity analysis to assess whether our conclusions depend on this outlier.

\textbf{Excluding granite-speech:} With N=6 Conformers, the architectural effects become \textit{stronger}, not weaker:
\begin{itemize}
    \item \textbf{Gender}: $\Delta p$ changes from $-0.126$ to $-0.154$; $p$ improves from 0.011 to 0.003, now surviving Bonferroni correction
    \item \textbf{Accent}: $\Delta p$ remains stable at $-0.23$; $p = 0.007$
    \item \textbf{Phoneme}: $\Delta p$ remains stable at $-0.29$; $p = 0.015$
\end{itemize}
This confirms that the ``Categorize Early'' hierarchy is a property of the Conformer architecture, not an artifact of the granite-speech outlier. Indeed, granite-speech's late Gender peak (0.32 vs. Conformer mean 0.13) was \textit{attenuating} the architectural effect.

\subsection{Quantifying the Multilingual Effect}
To isolate whether the ``Whisper Anomaly'' reflects multilingual training data rather than other factors, we performed a paired comparison of multilingual vs. English-only (.en) Whisper models at matched scales (tiny, base, small, medium; N=4 pairs).

\textbf{Results:} Surprisingly, multilingual training shows \textit{no significant effect} on peak positions:
\begin{itemize}
    \item Gender: $\Delta = -0.004$, $p = 0.93$ (no difference)
    \item Accent: $\Delta = -0.002$, $p = 0.97$ (no difference)
    \item Phoneme: $\Delta = -0.115$, $p = 0.30$ (trend, not significant)
    \item Duration: $\Delta = -0.007$, $p = 0.85$ (no difference)
\end{itemize}

\textbf{Interpretation:} The delayed gender and accent peaks in Whisper models are \textit{not} attributable to multilingual vs. monolingual training data. Instead, the ``Whisper effect'' likely reflects other factors shared across all Whisper models: (1) the sheer scale of training data (680K+ hours), (2) the multitask training objective (transcription + translation + language ID), or (3) architectural details specific to Whisper's implementation. This finding refines our earlier interpretation: the training \textit{objective} (multitask) rather than training \textit{data diversity} (multilingual) may be the key modulator of representational hierarchies.

\section{Feature Definitions}
\label{sec:appendix_features}
This appendix summarizes the feature taxonomy used for probing. Full dataset-specific extraction parameters, preprocessing, and label construction are included in Appendix \ref{sec:appendix_details}.
\begin{itemize}
    \item \textbf{Acoustic}: F0, formants (F1--F3), intensity, and related summary statistics.
    \item \textbf{Demographic}: speaker gender (binary) and L1 background / accent (categorical).
    \item \textbf{Phonetic}: phoneme identity (categorical).
    \item \textbf{Temporal}: utterance-level duration (continuous).
\end{itemize}

\section{Additional Results and Outliers}
\label{sec:appendix_additional_results}
We provide expanded discussion of representational-strength trends, entropy/concentration summaries, and model-family heterogeneity. This includes additional detail on models that depart from canonical hierarchies and on objective-driven modulation (e.g., multitask training), complementing the main-text ``hierarchy of controls'' framing.

\section{Outliers and Training-Regime Analyses}
\label{sec:appendix_outliers}
Several models illustrate within-family heterogeneity without changing the primary Transformer--Conformer separation. For example, Whisper models delay gender and accent peaks relative to SSL Transformers, consistent with objective-driven modulation. We also observe a small number of outliers (e.g., multimodal Phi-4; hubert-large; granite-speech) with atypical peak patterns or reduced probe strength. We retain these models to avoid selective exclusion and report robustness checks in Appendix \ref{sec:appendix_robustness}.

\section{Reproducibility}
\label{sec:appendix_repro}

All analyses were conducted using Python 3.9 with NumPy 2.0, SciPy 1.13, and scikit-learn 1.6. Probe training used PyTorch 2.0. Code for reproducing all analyses will be made publicly available upon publication. For the five architectural comparison t-tests (one per feature), we applied Bonferroni correction, yielding a corrected significance threshold of $\alpha = 0.05/5 = 0.01$.

\subsection{Software Versions}
\begin{itemize}
    \item Python 3.9
    \item NumPy 2.0.2
    \item Pandas 2.3.3
    \item SciPy 1.13.1
    \item Scikit-learn 1.6.1
    \item Statsmodels 0.14.6
    \item Matplotlib 3.9.2
    \item PyTorch 2.0
\end{itemize}

\subsection{Computational Resources}
Probing experiments were conducted on NVIDIA A100 GPUs. Each model's full layer-wise probing (all features, all layers) required approximately 2--8 hours depending on model size.

\subsection{Data Availability}
The L2-ARCTIC dataset is publicly available. Model weights are available from HuggingFace Hub under their respective licenses (Apache 2.0 for most models). Code and probing results will be released upon publication.

\end{document}